\documentclass[letterpaper]{article}
\usepackage{spconf,amsmath,graphicx,hyperref}
\usepackage{comment}
\usepackage{multicol}
\usepackage{makecell}
\usepackage{amssymb}
\usepackage{float}

\title{TS-Hint: Enhancing Semiconductor Time Series Regression Using Attention Hints From Large Language Model Reasoning}
\name{Jonathan Adam Rico$^{\star \dagger}$\thanks{The authors would like to thank SUTD PhD Scholarship for funding the research and A*STAR for providing computational resources.} \qquad Nagarajan Raghavan$^{\star}$ \qquad Senthilnath Jayavelu$^{\dagger}$}
  
  \address{$^{\star}$Engineering Product Development (EPD), SUTD, Singapore \\
      $^{\dagger}$Institute for Infocomm Research (I$^2$R), A*STAR, Singapore}

\begin{document}
%
\maketitle
\begin{abstract}
Existing data-driven methods rely on the extraction of static features from time series to approximate the material removal rate (MRR) of semiconductor manufacturing processes such as chemical mechanical polishing (CMP). However, this leads to a loss of temporal dynamics. Moreover, these methods require a large amount of data for effective training. In this paper, we
propose \textit{TS-Hint}, a Time Series Foundation Model (TSFM)
framework, integrated with chain-of-thought reasoning which
provides attention \textit{hints} during training based on attention mechanism data and saliency data. Experimental results demonstrate the effectiveness of our model in limited data settings via few-shot learning and can learn directly from multivariate time series features.
\end{abstract}
\begin{keywords}
Foundation model reasoning, Attention mechanism, Extrinsic regression, Semiconductor process, Multivariate time series
\end{keywords}
\section{Introduction}
\label{sec:intro}

With the increasing trend for the usage of computational electronic devices such as smartphones and computers, there is an increasing demand for semiconductor manufacturing processes. Among the critical processes of semiconductor manufacturing is chemical mechanical polishing (CMP) which requires planarization of the wafer thickness with precision in nanometers scale. However, the wafer thickness cannot be measured during the process and thus several heuristics and data-driven methods were used to approximate it and improve the efficiency of the polishing process. These methods range from traditional \cite{preston1927mrr, tseng1997mrr}, Physics-based \cite{yu2019mrr}, machine learning (ML) \cite{li2018}\cite{li2019}, and deep learning (DL) \cite{lim2021} models which rely heavily on techniques to extract static features from time series features.

There has been a recent trend in time series foundation models (TSFM) for downstream tasks. However, existing TSFMs do not have reasoning capabilities and most of them are applied to forecasting and classification, as per the author's knowledge there is no work on TSFM for extrinsic (sequence-to-one) regression task. TimeLLM \cite{jin2023timellm} incorporates large language model (LLM) reasoning but is limited to forecasting. Several advances demonstrated the success of LLM reasoning, such as chain-of-thought, for time series model integration \cite{wang2025_ts_cot, chow2024_ts_cot}. We take motivation from recent works \cite{selvaraju2019hint} that demonstrated the use of attention hint to improve the model performance in computer vision.

In this paper, we apply a novel TSFM with reasoning on semiconductor CMP process. TS-Hint achieved comparable performance with existing methods for the entire training dataset and demonstrated the benefits of the framework on limited data scenarios. The main contributions of this paper are summarized as follows:
\begin{itemize}
    \item We propose a multivariate (19 features) time series foundation model for extrinsic regression, able to learn directly from raw time series features, and demonstrate its effectiveness in a semiconductor dataset.
    \item We enhance the time series regression model with in-context learning via LLM chain-of-thought reasoning using transformer attention scores and saliency maps.
    \item We demonstrate the model's data efficiency with limited (15\%) training data and further few-shot fine-tuning is incorporated with LLM to understand the reason of the prediction.
\end{itemize}

\section{RELATED WORKS}
\label{sec:related_works}

The Deep Belief Network (DBN) model introduced in \cite{wang2017preston} has been reported as non-reproducible according to \cite{lim2021}. Nevertheless, \cite{wang2017preston} was the first to use the Preston equation as a benchmark to compare models for MRR prediction. The Physics-based approach proposed in \cite{yu2019mrr} uses a Random Forest (RF) model to determine the unknown variables in their equation. However, such empirical models require specific features, such as particle density and wafer hardness, which are not measured in experiments, leaving heuristics and substitute features for their equations. An ensemble of three tree-based models, Gradient-Boosted Trees (GBT), RF, and Extremely Randomized Trees (ERT), proposed in \cite{li2019} extracted statistical features such as central moment, skewness, kurtosis, and standard deviation. The feature engineering approach proposed in \cite{li2018} involves using features such as Stage, Wafer ID, Effective polishing time. However, these features are unconventional and include unique identifiers that may artificially correlate with the target MRR. The DL Autoencoder (AE) model proposed in \cite{lim2021} extracted statistical moments. However, these models depend on extracting static data from time series, resulting in loss of temporal information. 

Recent works have integrated LLM reasoning to enhance time series performance \cite{chen2024llmts, fan2025llmts} and some specifically used chain-of-thought (CoT) reasoning for time series \cite{wang2025_ts_cot, chow2024_ts_cot}. However, \cite{wang2025_ts_cot} and \cite{fan2025llmts} are limited to time series forecasting, while Chow et al. \cite{chow2024_ts_cot} only evaluated their framework on classification task and cannot be used as a benchmark for regression. LLM-TS Integrator \cite{chen2024llmts} proposed sample reweighting by LLM reasoning but is highly dependent on the quality of textual description of time series. On the other hand, our proposed framework integrates LLM chain-of-thought reasoning by suggesting attention hints, which demonstrate the benefits of attention-based TSFMs over ML and DL models. 

\section{METHODOLOGY}
\label{sec:method}

\subsection{Preliminaries}
Material removal rate (MRR) cannot be measured during the actual chemical mechanical polishing (CMP) process. It is defined as the reduction in the thickness of the semiconductor wafer over time given by

\begin{equation}
\label{eqn:mrr_actual}
    \text{MRR} = \frac{\Delta \text{Thickness}}{\Delta \text{Polishing Time}}
\end{equation}

Instead, the wafer thickness is measured before and after the process, where the duration of the polishing process (when to stop) is decided by heuristics and rules. The Preston equation \cite{preston1927mrr} provides one of the earliest heuristics to estimate the MRR of the CMP process. It is given by

\begin{equation}
\label{eqn:preston_mrr}
    \text{MRR} = kPV
\end{equation}
where $P$ is the downward pressure applied to the wafer, $V$ is the relative rotation speeds between the wafer and the polishing pad, and $k$ is a constant. 

\subsection{Time Series Foundation Model}

Time series foundation models (TSFM) have self-supervised capability, such that they can learn with minimal labeled samples and can be fine-tuned to several downstream tasks. Several TSFMs are limited to forecasting task \cite{jin2023timellm}, to univariate time series \cite{rasul2024lagllama, zhou2023gpt4ts}, and to sequence-to-sequence tasks \cite{garza2023timegpt,das2024timesfm, woo2024moirai}. However, there are no existing TSFM that can perform regression task. Among the TSFMs with classification capability, TimesNet \cite{wu2023timesnet} is not attention-based and Moment \cite{goswami2024moment} has complex architecture compared to PatchTST \cite{nie2023patchtst} which is the backbone of our framework for regression task. We adapt TSFMs with classification capability with the following main modifications: (i) replace the classification head with a multi-layer perceptron regression head and change the loss function to mean squared error during training and; (ii) standardize the input time series with fixed length by resampling the input sequence via linear interpolation.  PatchTST \cite{nie2023patchtst} converts time series into patches and passes to its transformer-based encoder where the attention score $A_h^{(i)}$ of sample $i$ at attention head $h$ can be retrieved using

\begin{equation}
\label{eqn:attention_score}
    A_h^{(i)} = \text{Softmax} \left( \frac{Q_h^{(i)} {K_h^{(i)}}^T}{\sqrt{d_k}} \right)
\end{equation}
where $Q, K,$ and $d_k$ are query, key, and attention head dimension respectively.  

\subsection{Chain-of-Thought Reasoning}

\begin{figure*}[tb]
    \centering
    \includegraphics[width=0.75\textwidth]{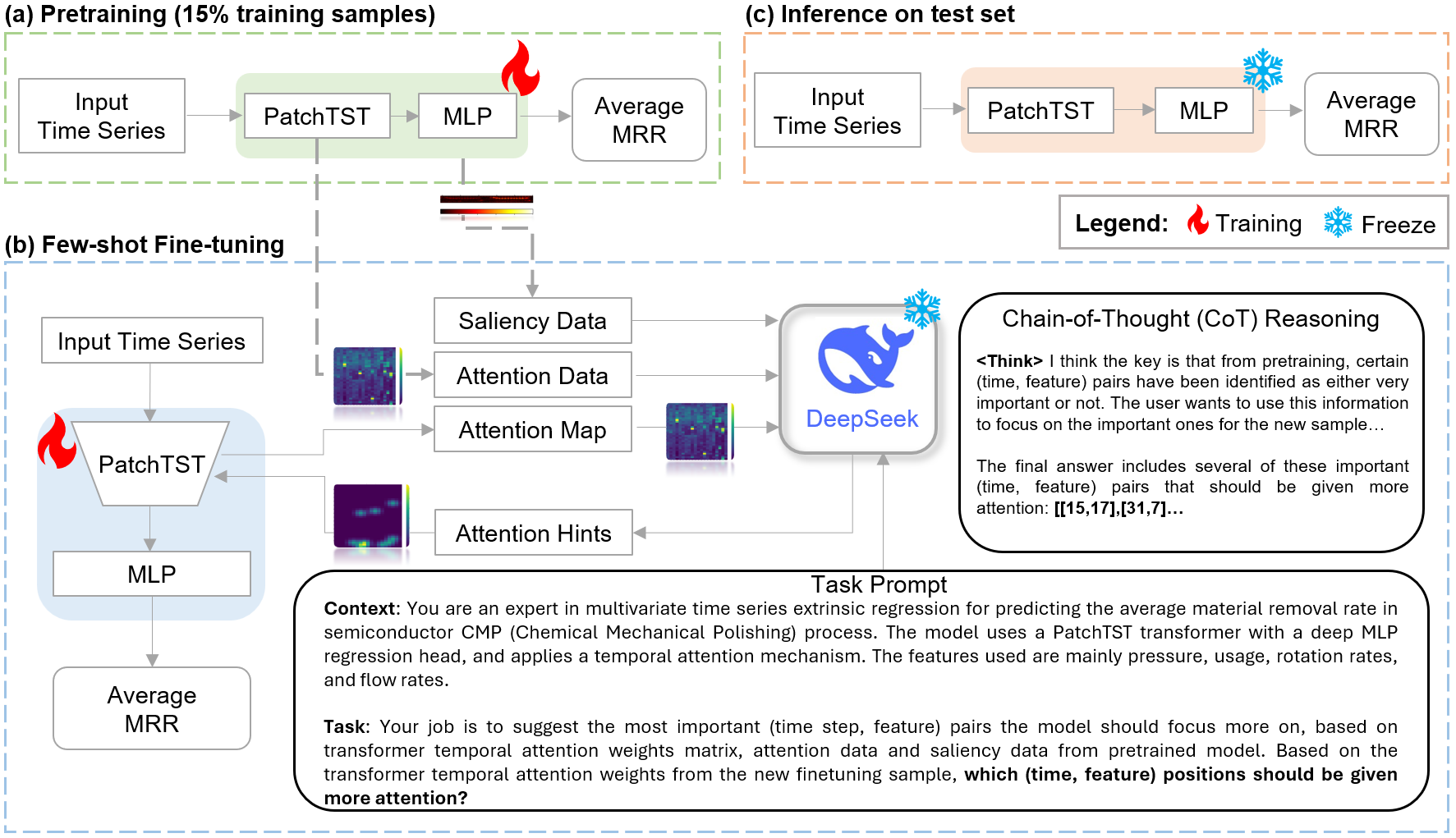}
    \caption{Overview of \textbf{TS-Hint architecture} shows (a) pretraining on 15\% training data, (b) few-shot fine-tuning one sample at a time enhanced by LLM chain-of-thought reasoning, and (c) inference on the test set. Dashed arrow lines indicate the attention data and saliency data retrieved from the pretrained model once.}
    \label{fig:model_architecture}
\end{figure*}

 Our proposed model, shown in Fig. \ref{fig:model_architecture}, is a multivariate TSFM for regression enhanced by the LLM chain-of-thought reasoning. Chain-of-thought in the form of [prompt, think, think,..., answer] improves the factuality of LLM reasoning by sequential thinking before answering, thereby reducing hallucination.

\textbf{Attention Maps}. Attention scores reveal where the model focuses, and by providing attention hints during training, the model learns to focus on more important parts of the input. It is similar to knowledge distillation from pretrained model to the model being fine-tuned. To obtain attention insights, we calculate the average attention map from the top $k$, here $k=5$, best samples from the pretrained model. The attention hint $H^{(i)}$ for sample $i$ is reshaped to that of the attention layer score so that the attention score $\hat{A}_h^{(i)}$ in Eq. \ref{eqn:attention_score} becomes

\begin{equation}
\label{eqn:attention_score_w_hint}
    \hat{A}_h^{(i)} = A_h^{(i)} + \lambda H^{(i)} 
\end{equation}
where $\lambda$ controls the model training sensitivity to attention hint $H$ at head $h$.

\textbf{Saliency Maps}. Since TSFM self-attention is calculated only per feature, due to channel independence, we cannot compare the attention scores between features. To obtain insights on feature importance, we calculate the average saliency of the top $k$, here $k=5$, best samples from pretrained model. The calculated saliency map $S$ reveals important features $C$ and timesteps $T$ by calculating the gradient of the output $f(x)$ with respect to the input $x$.

\begin{equation}
    S(x) = \left| \frac{\partial f(x)}{x} \right| \in \mathbb{R}^{C\times T}
\end{equation}

During fine-tuning, the model parameters are updated to minimize the loss function. The saliency maps were generated using the Saliency attribution method of the Captum library which implements a gradient-based approach \cite{simonyan2013saliency}. 

\section{EXPERIMENTS}
\label{sec:experiments}

\subsection{Dataset}
\begin{figure}[tb]
    \centering
    \includegraphics[width=0.23\textwidth]{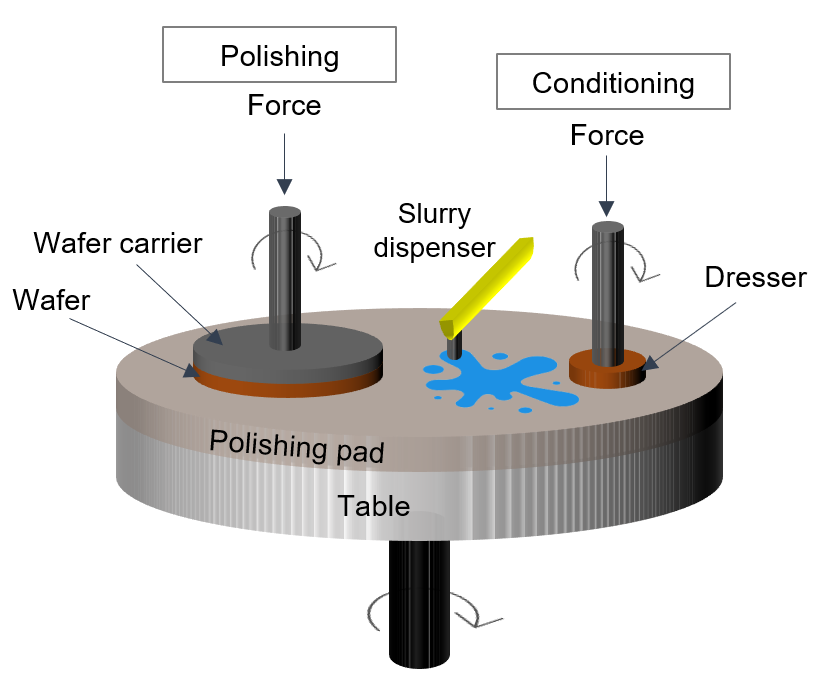}
    \caption{Chemical mechanical polishing setup.}
    \label{fig:cmp_setup}
\end{figure}

We benchmarked our model using the 2016 Prognostics and Health Management (PHM) dataset \cite{phm2016dataset} on semiconductor CMP process, for which the physical setup is shown in Fig. \ref{fig:cmp_setup}. The dataset includes static features such as Stage, Machine ID, Wafer ID, and Chamber ID but we only used the 19 time series features, including pressure, usage, rotation rates, and flow rates, as input to the TSFM.  Our model only uses the chamber ID to distinguish the mode for which the wafer sample was polished, as shown in Table \ref{tab:cmp_modes}. 

\begin{table}[tb]
    \centering
    \small
    \caption{The Chamber ID (static) provides the information of the mode of each CMP run.}
    \begin{tabular}{c|cccc}
        \Xhline{2\arrayrulewidth}
        \noalign{\vskip 2pt}
        Mode & Chamber & Average MRR & Train & Test\\
        \noalign{\vskip 2pt}
        \hline
        \noalign{\vskip 2pt}
        Low-speed & 4,5,6 &55-110 nm/min & 1613 & 364\\
        High-speed & 1,2,3 &140-170 nm/min & 351 & 73\\
        \noalign{\vskip 2pt}
        \Xhline{2\arrayrulewidth}
    \end{tabular}
    \label{tab:cmp_modes}
\end{table}

\subsection{Performance Evaluation}
 \textbf{Experimental Settings}. We ran the numerical experiments on a computer workstation with a NVIDIA 16 GB GPU GeForce RTX 4080 SUPER. DeepSeek-R1 14B via Ollama takes up to 12 GB in GPU memory during reasoning. The regression head has six hidden layers [1024,512,256,128,64,32] and ReLU activation functions. For training, we used Adam Optimizer with an initial learning rate of $0.001$ and MSE loss function. For fine-tuning, the learning rate is reduced to $0.0001$.
 
\begin{figure}[tb]
    \centering
    \includegraphics[width=0.25\textwidth]{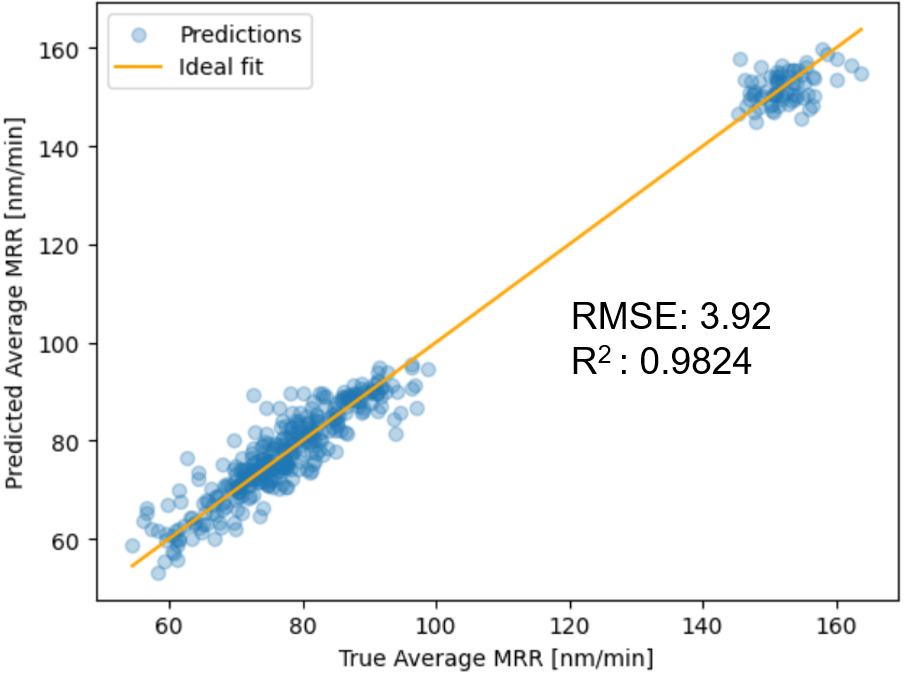}
    \caption{Scatter plot of true vs predicted average MRR of the TS-Hint for regression using full train data.}
    \label{fig:regression_fit}
\end{figure}

 Table \ref{tab:few_shot_results} demonstrates the effectiveness of our framework in limited data settings. Since there is no pretrained TSFM for the regression task, we pretrain using 15\% training samples. Starting with the pretrained (0-shot) performance 8.86 RMSE, the model improves the regression performance with few-shot (5-shot) learning down to 8.20 RMSE. To make par comparison with existing methods, we also ran the proposed TS-Hint framework using full training data and compared with existing methods such as Preston \cite{wang2017preston}, Physics-based ML \cite{yu2019mrr}, GBDT \cite{li2019}, RF \cite{li2018}, Autoencoder \cite{lim2021}, TimesNet \cite{wu2023timesnet}, and Moment \cite{goswami2024moment}. We evaluated the models using common regression metrics such as the root mean squared error (RMSE) and the coefficient of determination (R$^2$). Table \ref{tab:benchmark_models} shows that TS-Hint with 3.92 RMSE outperformed all benchmark models except for the Random Forest \cite{li2018} which achieved 2.72 RMSE. We tried reproducing \cite{li2018} and achieved 2.98 RMSE. However, removing the "STAGE" static feature, degrades the performance to 3.57 RMSE. TS-Hint can be competitive in multivariate time series extrinsic regression against ML and DL models only by using time series features. Fig. \ref{fig:regression_fit} shows the scatter plot of the regression fit illustrating that the model tends to overpredict the target.  

\begin{table}[tb]
    \centering
    \small
    \caption{Few-shot fine-tuning performance on the test set.}
    \begin{tabular}{c|cc}
        \Xhline{2\arrayrulewidth} 
        N-shot Learning & RMSE & R$^2$ \\
        \hline
        0-shot & 8.8678 & 0.9100\\
        1-shot & 8.6903 & 0.9135\\
        2-shot & 8.9478 & 0.9083\\
        3-shot & 8.6956 & 0.9134\\
        4-shot & 8.4221 & 0.9188\\
        5-shot & \textbf{8.2022} & \textbf{0.9230}\\
        \Xhline{2\arrayrulewidth} 
    \end{tabular}   
    \label{tab:few_shot_results}
\end{table}

\begin{table}[tb]
    \centering
    \small
    \caption{RMSE scores of benchmark models on the test set using full training data.}
    \begin{tabular}{l|l|c}
        \Xhline{2\arrayrulewidth} 
        Type & Model & RMSE \\
        \hline
        Traditional, & Preston & 29.5 \\
        Hybrid & Phys-inf ML & 15.77 \\
        \hline
        ML & GBDT & 4.64 \\
           & RF & \textbf{2.72} \\
        \hline
        DL & AE & 4.77 \\
        \hline
             & Moment & 6.17 \\
        TSFM & TimesNet & 4.21 \\
             & \textbf{TS-Hint (Ours)}& \textbf{3.92} \\
        \Xhline{2\arrayrulewidth} 
    \end{tabular}
    \label{tab:benchmark_models}
\end{table}

\begin{figure}[tb]
    \centering
    \includegraphics[width=0.5\textwidth]{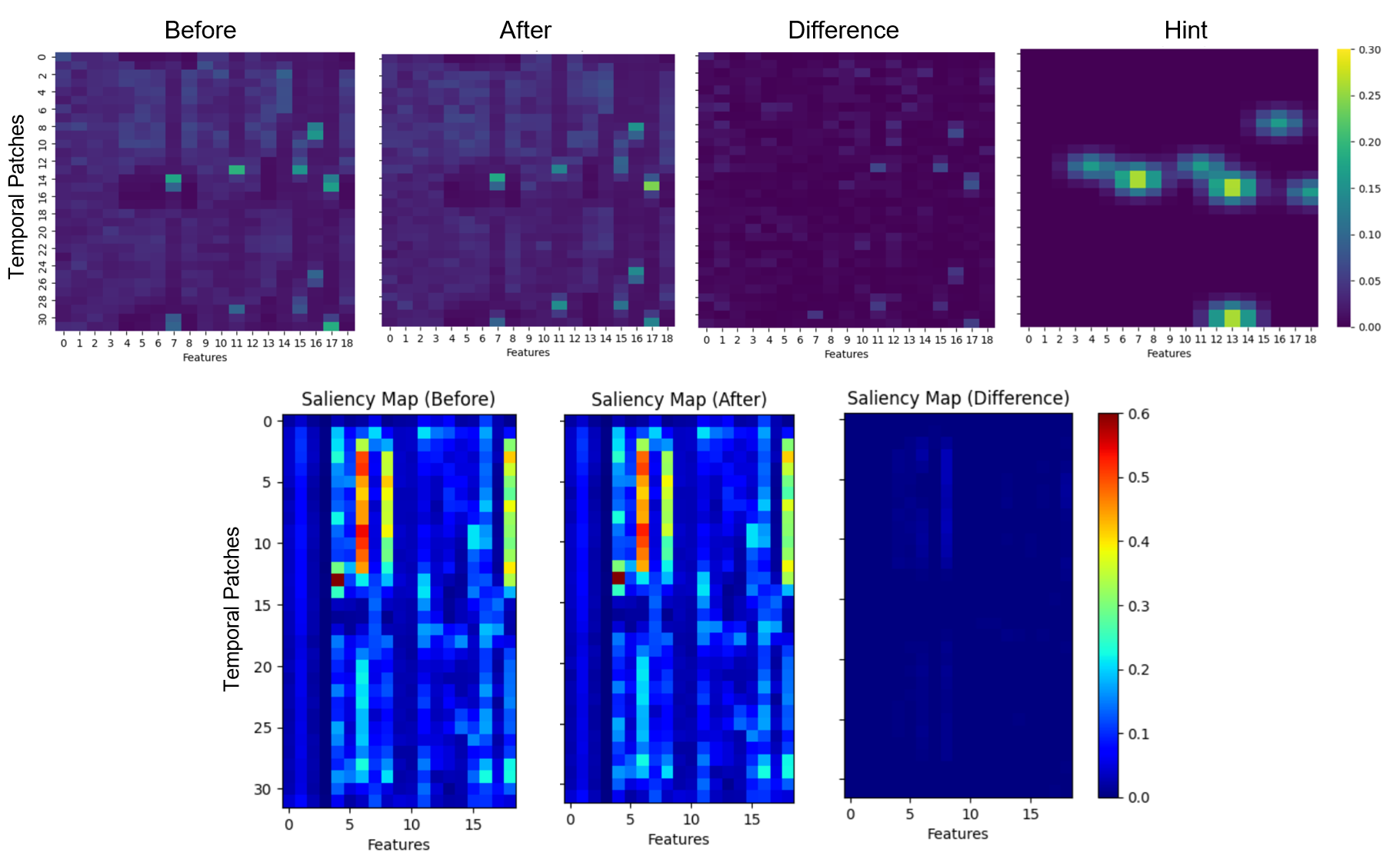}
    \caption{Before, after, and absolute difference for 1-shot fine-tuning (top) attention maps and attention hint and (bottom) saliency maps.}
    \label{fig:attention_hints}
\end{figure}

\textbf{Reasoning Effectiveness}. Fig. \ref{fig:attention_hints} shows the attention maps before and after 1-shot fine-tuning, the attention hint suggested by the LLM chain-of-thought reasoning, and the saliency maps before and after 1-shot fine-tuning. The absolute differences between the before and after are also shown in Fig \ref{fig:attention_hints} which illustrates where the adjustments in attention and gradient attribution occurred. For example, the attention shifted to features 15-17, Stage rotation, Head rotation, and Dressing water status, and the gradient attribution is slightly reduced in feature 8, Ripple air bag pressure. The model's reasoning is backed by data-driven insights based on what has worked well in the past and what is currently important. For example, the LLM considers a handpicked combination of the important timesteps from the given attention insights and the important features from the given saliency insights. Incorporating the attention hint during few-shot fine-tuning results in a decrease in training loss in most cases. For example, after 1-shot fine-tuning, the training loss decreased from 0.7710 to 0.6085. This improvement did not result in overfitting the train set as the model performance also improved in the test set shown in Table \ref{tab:few_shot_results}. 

\section{CONCLUSION}
\label{sec:conclusion}

We developed a multivariate time series foundation model enchanced by LLM chain-of-thought reasoning. We demonstrated the effectiveness of our framework on limited data scenarios via a few-shot fine-tuning. Our model achieved 3.92 RMSE which is comparable to the top regression models in CMP PHM dataset showing the model's ability to learn directly from time series data. We hope that our work provides a stepping stone for LLM reasoning integration utilizing attention scores and saliency weights. Despite this, we recognize that LLMs still struggle to understand time series data and high-dimensional matrix since LLMs reason more symbolically than numerically. Unfortunately, integrating LLM reasoning during training makes training process much longer and computationally expensive. For future work, we plan to extend LLM reasoning with self-reflection reasoning and letting the LLM tune certain parameters (number of hints, hint weights, and hint smoothen kernel size) when applying attention hints. 


\clearpage
\bibliographystyle{IEEEbib}
\bibliography{strings,refs}

@inproceedings{goswami2024moment,
  title={MOMENT: A Family of Open Time-series Foundation Models},
  author={M. Goswami and K. Szafer and A. Choudhry and Y. Cai and S. Li and A. Dubrawski},
  booktitle={ICML},
  year={2024}
}

@inproceedings{zhou2023gpt4ts,
  title={{One Fits All}: Power General Time Series Analysis by Pretrained LM},
  author={T. Zhou and P. Niu and X. Wang and L. Sun and R. Jin},
  booktitle={NeurIPS},
  year={2023}
}

@misc{rasul2024lagllama,
      title={Lag-Llama: Towards Foundation Models for Probabilistic Time Series Forecasting}, 
      author={K. Rasul and A. Ashok and A. R. Williams and H. Ghonia and R. Bhagwatkar and A. Khorasani and M. J. D. Bayazi and G. Adamopoulos and R. Riachi and N. Hassen and M. Biloš and S. Garg and A. Schneider and N. Chapados and A. Drouin and V. Zantedeschi and Y. Nevmyvaka and I. Rish},
      year={2024},
      eprint={2310.08278},
      archivePrefix={arXiv},
      primaryClass={cs.LG}
}

@inproceedings{jin2023timellm,
  title={{Time-LLM}: Time series forecasting by reprogramming large language models},
  author={Jin, M. and Wang, S. and Ma, L. and Chu, Z. and Zhang, J. Y. and Shi, X. and Chen, P. and Liang, Y. and Li, Y.F. and Pan, S. and Wen, Q.},
  booktitle={ICLR},
  year={2024}
}

@inproceedings{wu2023timesnet,
  title={TimesNet: Temporal 2D-Variation Modeling for General Time Series Analysis},
  author={H. Wu and T. Hu and Y. Liu and H. Zhou and J. Wang and M. Long},
  booktitle={ICLR},
  year={2023},
}

@article{garza2023timegpt,
    author = {A. Garza and M. Mergenthaler-Canseco},
    title = {TimeGPT-1},
    journal = {arXiv},
    year = {2023}
}

@article{das2024timesfm,
    author = {Das, A. and W. Kong and R. Sen and Y. Zhou},
    title = {A decoder-only foundation model for time-series forecasting},
    journal = {ICML},
    year = {2024}
}

@inproceedings{woo2024moirai,
  title        = {Unified Training of Universal Time Series Forecasting Transformers},
  author       = {Woo, G. and Liu, C. and Kumar, A. and Xiong, C. and Savarese, S. and Sahoo, D.},
  booktitle    = {Proceedings of the 41st ICML},
  series       = {PMLR},
  volume       = {235},
  pages        = {53140-53164},
  year         = {2024},
  url          = {https://proceedings.mlr.press/v235/woo24a.html}
}

@article{nie2023patchtst,
    author = {Nie, Y. and Nguyen, N.H. and Sinthong, P. and Kalagnanam, J.},
    title = {A time series is worth 64 words: Long-term forecasting with transformers},
    journal = {ICLR},
    year = {2023}
}

@article{preston1927mrr,
    author = {Preston, F.W.},
    title = {The theory and design of plate glass polishing machines},
    journal = {J. Society of glass Tech},
    year = {1927}
}

@article{tseng1997mrr,
    author = {Tseng, W.T. and Wang, Y.L.},
    title = {Re‐examination of pressure and speed dependences of removal rate during chemical‐mechanical polishing processes},
    journal = {Journal of the Electrochemical Society},
    year = {1997},
    page = {15},
    volume = {144},
    issue = {2}
}

@article{li2019,
    author = {Li, Z. and Wu, D. and Yu, T},
    title = {Prediction of material removal rate for chemical mechanical planarization using decision tree-based ensemble learning},
    journal = {Journal of Manufacturing Science and Engineering},
    year = {2019},
    page = {031003},
    volume = {141},
    issue = {3}
}

@inproceedings{lim2021,
    author = {Lim, K.L. and Dutta, R.},
    title = {Prognostics and health management of wafer chemical-mechanical polishing system using autoencoder},
    booktitle = {Proceedings ICPHM},
    organization = {IEEE},
    year = {2021},
    pages = {1-8},
}

@article{yu2019mrr,
    author = {Yu, T. and Li, Z. and Wu, D.},
    title = {Predictive modeling of material removal rate in chemical mechanical planarization with physics-informed machine learning},
    journal = {Wear},
    year = {2019},
    volume = {426},
    pages = {1430-1438}
}

@article{li2018,
    author = {Li, X. and Wang, C. and Zhang, L. and Mo, X. and Zhao, D. and Li, C.},
    title = {Assessment of physics-based and data-driven models for material removal rate prediction in chemical mechanical polishing},
    journal = {2nd ICEEA},
    year = {2018}
}

@article{wang2017preston,
    author = {Wang, P. and Gao, R.X. and Yan, R.},
    title = {A deep learning-based approach to material removal rate prediction in polishing},
    journal = {CIRP annals},
    year = {2017},
    volume = {66},
    issue = {1},
    pages = {429-432}
}

@article{chen2024llmts,
    author = {Chen, C. and Oliveira, G. and Noghabi, H.S. and Sylvain, T.},
    title = {LLM-TS Integrator: Integrating LLM for Enhanced Time Series Modeling},
    journal = {arXiv preprint arXiv:2410.16489},
    year = {2024}
}

@article{fan2025llmts,
    author = {Fan, H. and Li, B. and Weng, Y. and Zhou, S.},
    title = {Small but mighty: enhancing time series forecasting with lightweight LLMs},
    journal = {The Journal of Supercomputing},
    year = {2025},
    volume = {81},
    issue = {8},
    pages = {985}
}

@article{wang2025_ts_cot,
    author = {Wang, X. and Zhou, T. and Gao, J. and Ding, B. and Zhou, J.},
    title = {Output Scaling: YingLong-Delayed Chain of Thought in a Large Pretrained Time Series Forecasting Model},
    journal = {preprint arXiv:2506.11029},
    year = {2025}
}

@inproceedings{chow2024_ts_cot,
  title={Towards Time-Series Reasoning with LLMs},
  author={Chow, W. and Gardiner, L. and Hallgrimsson, H. and Xu, M. and Ren, S.},
  booktitle={Adv. in NeurIPS},
  volume={37},
  year={2024},
  publisher={NeurIPS Foundation},
  url={https://neurips.cc/virtual/2024/102927}
}

@inproceedings{selvaraju2019hint,
    author = {Selvaraju, R.R. and Lee, S. and Shen, Y. and Jin, H. and Ghosh, S. and Heck, L. and Batra, D. and Parikh, D.},
    title = {Taking a hint: Leveraging explanations to make vision and language models more grounded},
    booktitle = {Proceedings ICCV},
    organization = {IEEE},
    year = {2019},
    pages = {2591-2600}
}

@article{simonyan2013saliency,
  title={Deep Inside Convolutional Networks: Visualising Image Classification Models and Saliency Maps},
  author={K. Simonyan and A. Vedaldi and A. Zisserman},
  journal={ICLR},
  year={2013},
  volume={abs/1312.6034},
}

@article{phm2016dataset,
    author = {Society of Prognostic and Health Management},
    title = {PHM Data Challenge},
    journal = {},
    note = {\url{http://www.phmsociety.org/events/conference/phm/16/data-challenge.}},
    year = {2016}
}

\end{document}